\documentclass[sigconf]{acmart}

\AtBeginDocument{
  }
\usepackage{graphicx}
\usepackage{subfigure}
\usepackage{xspace}
\usepackage{stfloats}
\usepackage{array, multirow}
\usepackage{amsmath,amsfonts}
\usepackage{algorithm,algorithmic}
\usepackage{textcomp}
\usepackage{xcolor}
\usepackage{tikz}
\usepackage{makecell}
\usepackage{bm}
\usepackage{balance}

\usepackage{xspace}

\newcommand{\camera}[1]{{\color{black}{#1}}}

\copyrightyear{2024}
\acmYear{2024}
\setcopyright{rightsretained}
\acmConference[E-Energy '24]{The 15th ACM International Conference on Future
and Sustainable Energy Systems}{June 4--7, 2024}{Singapore, Singapore}
\acmBooktitle{The 15th ACM International Conference on Future and Sustainable
Energy Systems (E-Energy '24), June 4--7, 2024, Singapore,
Singapore}
\acmDOI{10.1145/3632775.3661962}
\acmISBN{979-8-4007-0480-2/24/06}

\begin{document}

\title{A Dataset for Research on Water Sustainability}
\author{Pranjol Sen Gupta}
\orcid{0000-0002-0146-9148}
\affiliation{
  \institution{The University of Texas at Arlington}
  \country{}
}
\email{pranjolsen.gupta@mavs.uta.edu}

\author{Md Rajib Hossen}
\orcid{0000-0002-0882-2434}
\affiliation{
  \institution{The University of Texas at Arlington}
  \country{}
}
\email{mdrajib.hossen@mavs.uta.edu}

\author{Pengfei Li}
\orcid{0000-0003-3257-9929}
\affiliation{
  \institution{University of California,
Riverside}
\country{}
}
\email{pli081@ucr.edu}

\author{Shaolei Ren}
\orcid{0000-0001-9003-4324}
\affiliation{
  \institution{University of California,
Riverside}
\country{}
}
\email{sren@ece.ucr.edu}
\author{Mohammad A. Islam}
\orcid{0000-0002-5778-4366}
\affiliation{
  \institution{The University of Texas at Arlington}
  \country{}
}
\email{mislam@uta.edu}
\begin{abstract}
Freshwater scarcity is a global problem that requires collective efforts across all industry sectors. Nevertheless, a lack of access to operational water footprint data bars many applications from exploring optimization opportunities hidden within the temporal and spatial variations. To break this barrier into research in water sustainability, we build a dataset for operation direct water usage in the cooling systems and indirect water embedded in electricity generation. Our dataset consists of the hourly water efficiency of major U.S. cities and states from 2019 to 2023. We also offer cooling system models that capture the impact of weather on water efficiency. We present a preliminary analysis of our dataset and discuss three potential applications that can benefit from it.
Our dataset is publicly available at Open Science Framework (OSF) \cite{osf_water_dataset}.
\end{abstract}

\ccsdesc[500]{General and reference~Metrics}
\ccsdesc[500]{General and reference~Measurement}
\ccsdesc[500]{Hardware~Impact on the environment}
\keywords{Sustainability, Water Consumption, Dataset, Cooling Tower, Electricity Generation}

\maketitle
\section{Introduction}

\begin{figure}[t]
	\vspace{-0.0cm}
	\centering 
    \includegraphics[width=1\linewidth]{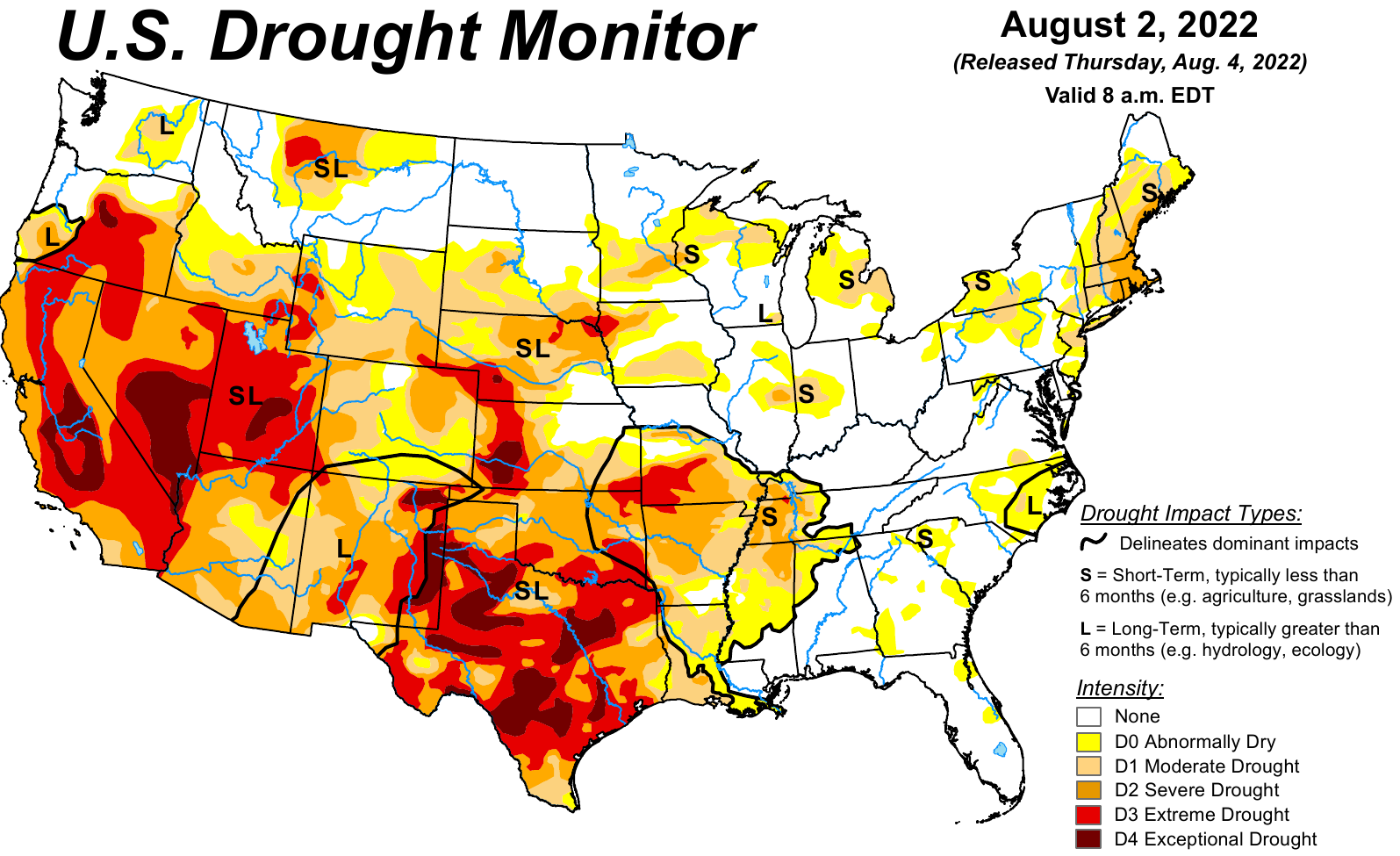}
\caption{US drought map for August 2, 2022, with 4.47\% area under exceptional drought (D4), 18.96\% area under extreme drought or worse (D3-D4), and 37.03\% area under severe drought or worse (D2-D4) \cite{usDroughtMap}} \label{fig:drought_map}
	\vspace{-0.0cm}
\end{figure}

Global freshwater supply is under immense pressure due to the growing population and deteriorating climate conditions, making extended droughts a norm in many parts of the world \cite{Water_Shortage_2030}. For example,
Fig.~\ref{fig:drought_map} shows that 37.03\% of the U.S. area was under severe drought or worse in 2022 \cite{usDroughtMap}.
Even in regions not historically prone to drought, such as the eastern U.S., the importance of water conservation persists due to aging public water infrastructure \cite{Water_Shortage_FL}. Therefore, every industry and application sector must scrutinize its water footprint and actively contribute to water sustainability efforts \cite{epa_water_sense, Google_SustainabilityReport_2023,10417786}.

In this paper, we focus on two pervasive components that significantly contribute to the water footprint across many sectors --- the water consumption in the cooling system and the water footprint embedded in the electricity consumption.
Evaporative cooling has been extensively used for the temperature regulation of buildings, and the water usage in these cooling systems can account for more than 50\% of the building's total use \cite{water_in_building_1}.
Meanwhile, the water footprint in electricity generation remains high despite a steady gain in water efficiency over the years. In 2021, every megawatt-hour of electricity generation used nearly 12,000 gallons of water \cite{water_in_electricity}.

The substantial water footprints in cooling systems and electricity consumption also present considerable opportunities for savings.
More importantly, both of these water footprints under scrutiny vary over time --- the cooling system's water consumption varies with local weather conditions, whereas the water embedded in electricity changes with variations in electricity generation sources. 
These temporal variations can be leveraged by applications like EV charging, which benefit from scheduling flexibilities \cite{alinia2019online}.  
Moreover, large cloud-scale data center applications can also incorporate in their load balancing the variations in water footprint across geographical locations \cite{liu2014greening,islam2016exploiting}.
However, the general lack of access to operational water footprint data hinders the development of water-sustainable operation strategies across many applications, such as building management and EV charging.

To foster water sustainability research and open up the largely untapped optimization opportunities in the temporal and spatial variations of water footprints, we build a water efficiency dataset that provides the hourly water efficiency of the cooling system and electricity generation across major U.S. cities and states \camera{from January 2019 to December 2023}. We also present our cooling system models to capture the impact of weather conditions on the water consumption rate of the cooling system.
\camera{Our dataset, along with all source data, models, and scripts, are made publicly available at 
Open Science Framework (OSF) \cite{osf_water_dataset}
}.
We conduct a preliminary analysis of our dataset, offering insights into temporal and spatial variations of water efficiency. 
Additionally, we discuss three sample applications that can benefit from our dataset.

\section{Preliminaries}

\begin{table*}[t!]
\caption{Estimated water intensity of different electricity generation sources in the U.S. \cite{Water_Electricity_EWIF_Water_Intensity_WorkingPaper_WorldResourcesInstitute_2020_reig2020guidance}.}
\label{tab:wue_fuel}
\begin{tabular}{|l|l|l|l|l|l|l|l|l|}
\hline
\textbf{Energy Source}             & Coal  & Hydro  & Natural Gas & Nuclear & Other & Petroleum & Solar & Wind  \\ \hline
\textbf{Water Intensity} (L/kWh) & 1.817 & 22.675 & 0.795       & 2.309   & 0.757 & 1.363     & 0.000 & 0.000 \\ \hline
\end{tabular}
\end{table*}

\subsection{Water Footprint}

\textbf{Withdrawal vs consumption.}
These are two important terms in the context of water management that warrant clarification \cite{Water_Consumption_Withdrawal_WorldResourceInstitute,Water_EWIF_macknick2011review}.
Water withdrawal is the total amount of water taken from a water source. It does not imply permanent removal and can be returned to the source after use, although it may undergo changes in quality or temperature during the process. Water consumption refers to the portion of withdrawn water that is not returned to its source. It represents the amount of water that is either evaporated, incorporated into products, or otherwise not available for immediate reuse in the same water source. 
While water withdrawal is integral for sustainable water use \cite{Water_Consumption_Withdrawal_WorldResourceInstitute}, our focus in this work centers on water consumption, which poses a more imminent threat to available water
and also consistent with the water footprint literature \cite{Water_EWIF_macknick2011review}.

\textbf{Direct water consumption.}
It refers to the water, an entity consumes for its own operational processes and activities. It involves water that is physically used on-site and is often directly under the control or management of the entity.
Examples of direct water consumption are water consumed in the cooling systems, sanitation and cleaning, irrigation, and fire suppression.

\textbf{Indirect water consumption.}
It refers to the water consumed beyond an entity's operational boundaries and direct control. Examples of an entity's indirect water consumption are water usage in the production of energy consumed, supply chain, manufacturing and transportation equipment, and consumer use. 

To enable holistic sustainability, we can align the direct and indirect water footprint with the Greenhouse Gas Protocol's (GHG Protocol) widely accepted accounting approach for measuring and managing greenhouse gas emissions \cite{protocol2011greenhouse}.
While the GHG Protocol has been primarily associated with greenhouse gas emissions, it can be extended to include water usage as well \cite{Water_Electricity_EWIF_Water_Intensity_WorkingPaper_WorldResourcesInstitute_2020_reig2020guidance}.
Following the GHG Protocol, the direct water footprint can be considered Scope 1, while the indirect water footprint falls under Scopes 2 and 3.

\subsection{Water Consumption in Cooling Systems}
Commercial high-capacity cooling systems, including those
used in office buildings and many data centers, typically use water in their cooling system for heat transport and dissipation into the environment \cite{Google_Water}.  
As illustrated in Fig.~\ref{fig:water_footprint}, these cooling systems consist of two water loops --- the inner loop carries the heat from the facility air handlers to the chiller heat exchanger, and the outer loop carries the heat from the chiller heat exchanger to the cooling tower for releasing the heat. The inner loop is closed and does not lose any water. The outer loop, on the other hand, sends hot water to the cooling tower, which cools down the water using water evaporation and, therefore, loses water in the process. This water loss through evaporation is the direct water consumption for the cooling system. Note that the water in the outer loop also requires regular recycling (known as blowdown) to avoid any buildup from concentrated minerals due to evaporation. The water lost through blowdown, however, is not considered water consumption as it is returned to the source as grey water.

\subsection{Water Consumption in Electricity Generation}

Different electricity generation sources have varying water consumption patterns, and their impact on water resources depends on the technology and processes involved \cite{macknick2012operational,spang2014water}.
Thermal power plants such as coal, natural gas, and oil use water for cooling purposes in the generation process. Nuclear power plants also require water for cooling, typically through cooling towers or direct discharges. Solar and wind power generation technologies have minimal water consumption during the electricity generation phase. Water use is mostly associated with the manufacturing and maintenance of the equipment.
Hydropower plants' water consumption, on the other hand, mainly comes from expedited surface evaporation in their water reservoirs.
Table~\ref{tab:wue_fuel} shows the U.S. average water consumption to generate a kilowatt-hour of electricity \cite{Water_Electricity_EWIF_Water_Intensity_WorkingPaper_WorldResourcesInstitute_2020_reig2020guidance}.

\subsection{Water Usage Effectiveness (WUE)}

Water Usage Effectiveness (WUE) is an operational water metric that quantifies the water efficiency of a system, as defined below
\begin{align}
    \text{WUE} = \frac{\text{Water Consumption}}{\text{Energy Processed/Generated}}
\end{align}
For a cooling system, WUE is the ratio of water consumption to the amount of heat dissipated, whereas, for electricity, it is the ratio of water consumption and electric energy generation.
\section{Water Efficiency Dataset}

\begin{figure}
    \centering
    \includegraphics[width=1\linewidth]{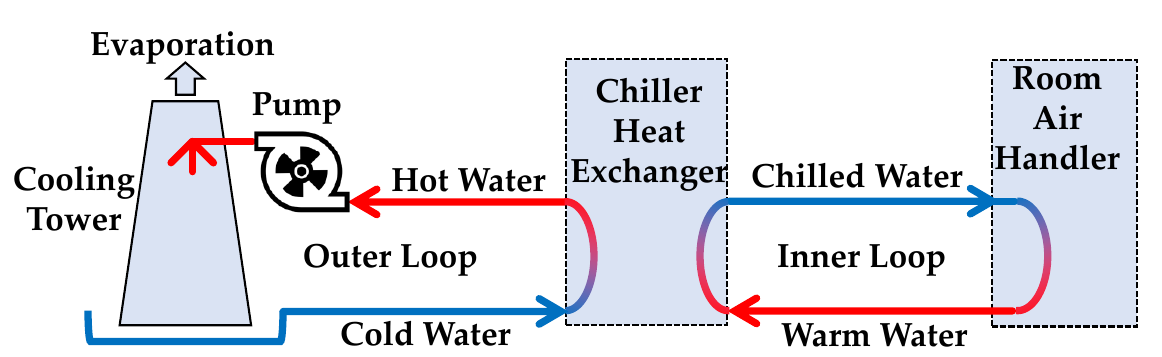}
    \caption{Cooling system with evaporative cooling.}
    \label{fig:water_footprint}
\end{figure}

\subsection{Methodology}

\textbf{Estimation of direct WUE.}
A cooling tower's water consumption varies with the outside air temperature and humidity. More specifically, the cooling tower consumes more water (i.e., water evaporates away) for the same cooling load when the outside air temperature is higher or more humid. 
While the precise relationship between WUE and weather conditions for a specific cooling system can vary, we here offer a generic model that captures the impact of the weather conditions on commercial cooling towers. We utilize the Water Calculator tool offered by SPX Technologies \cite{CoolingTower_SPX} to derive our model. 

\camera{
Before introducing our model, we briefly discuss the operational set points
that are crucial to understanding cooling towers and the SPX water calculator, as illustrated in Fig.~\ref{fig:water_footprint}.
The ``range'' denotes the temperature difference between hot and cold water in the outer loop. The range and flow rate determine the system's cooling load.
SPX water calculator captures the impact of weather conditions by including the wet bulb temperature. 
Wet bulb temperature indicates the temperature of air saturated with water and can be measured from a wet cloth exposed to airflow.
Wet bulb temperature can be measured from a wet cloth exposed to airflow.
For cooling tower operation, there must be a temperature difference between the wet-bulb temperature and the cold water temperature. This temperature difference is called the ``approach''.

Next, to analyze the impact of weather conditions on cooling tower water consumption using the SPX water calculator, first we set the flow rate to 1000 gallons per minute and a range of 10 Fahrenheit.
Note that the flow rate set point here is chosen to be an arbitrary round number and does not affect the WUE.
This configuration results in a cooling capacity of 1466 kW or 417 tons.
Meanwhile, the drift rate and concentration are maintained at their SPX default values of 0.005\% and 3, respectively. The drift rate refers to water droplets carried away from the cooling tower by airflow, while concentration indicates how many times water circulates in the outer loop before being discarded. In the SPX water calculator, drift rate and concentration values do not influence the water evaporation rate.
}

\begin{figure}
    \centering
\subfigure[]{ \label{fig:water_efficiency}    \includegraphics[width=0.48\linewidth]{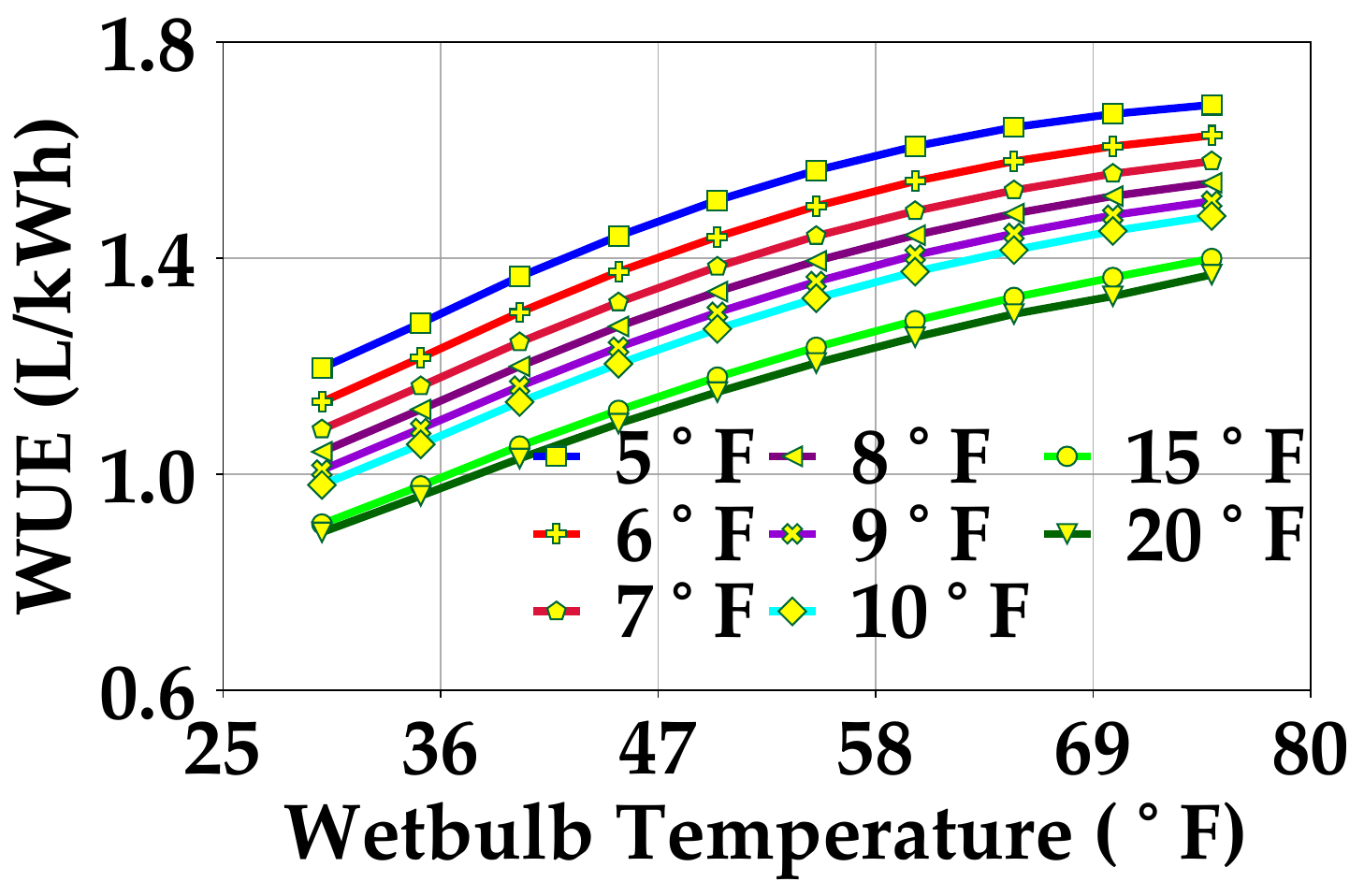}}
\subfigure[]{ \label{fig:water_efficiency_model}    \includegraphics[width=0.48\linewidth]{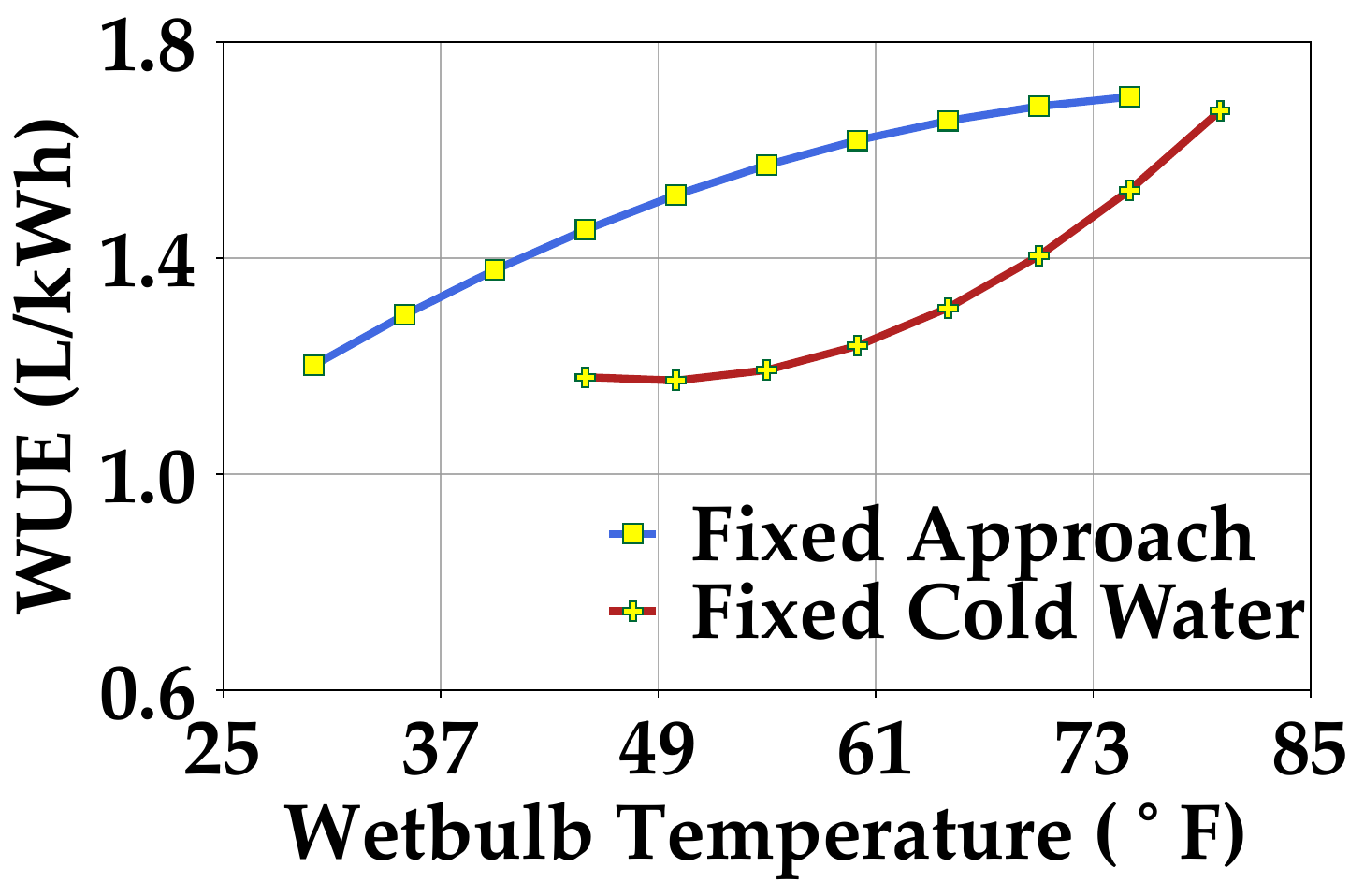}}
    \caption{(a) Change in WUE with wet bulb temperatures with different approaches (i.e., the difference between cold water temperature and wet bulb temperature). (b) Model of weather impact on direct WUE. For the "Fixed Approach" model, the approach is set at 5$^{\circ}$F. For the "Fixed Cold Water" temperature model, cold water temperature is set at 85$^{\circ}$F.}
\end{figure}

Fig.~\ref{fig:water_efficiency} shows the water efficiency of the cooling tower at different wet-bulb temperatures and different approaches. 
We see that at any given wet-bulb temperature, the WUE goes up with a decreasing approach. While this indicates a higher approach will result in lower water consumption, it also leads to a higher temperature of the cold water in the chiller heat exchanger, requiring the heat exchanger to work harder (and consume more energy) to transfer the heat from the inner loop to the outer loop. 
Here, we offer two different models, considering two operation strategies. In the first one, we \emph{fix the approach} to 5$^{\circ}$F to maximize the heat exchanger efficiency. In the second one, we \emph{fix the cold water temperature} to 85$^{\circ}$F (the typical maximum allowed temperature for a heat exchanger) and, therefore have a variable approach with changing wet bulb temperature.
Using these two strategies on our data points in Fig.~\ref{fig:water_efficiency_model}, we derive the following two models
\begin{align}\label{eqn:wue_fixedapproach}
    &W_{direct}^{FixedApproach} = - 0.0001896\cdot T_{w}^2+0.03095 \cdot T_{w}+0.4442 \\
    \label{eqn:wue_fixedcold}
    & W_{direct}^{FixedColdWater} = 0.0005112\cdot T_w^2 - 0.04982\cdot T_w + 2.387   
\end{align}
where $T_w$ is the wet bulb temperature in Fahrenheit.
\camera{Note here that, due to incompatible operation set points in cold weather conditions, in Eqns.~\eqref{eqn:wue_fixedapproach} and \eqref{eqn:wue_fixedcold}, the lower limits for $T_w$ are 30$^{\circ}$F and 45$^{\circ}$F, respectively.}
We show our models in Fig.~\ref{fig:water_efficiency_model}.

\camera{
In deriving Eqns.~\eqref{eqn:wue_fixedapproach} and \eqref{eqn:wue_fixedcold} above, we consider 100\% heat-transfer efficiency at the chiller heat exchanger and perfect thermal isolation in the water loops.
That is, the cooling load is solely determined by the flow rate and temperature difference between hot and cold water.
Moreover, we use SPX's own estimation of water evaporation rates in their cooling towers.

In practice, the cooling system efficiency can vary depending on the specific installation and manufacturer. 
This may increase or decrease the tower's water evaporation rate. 
To capture such variability, we can introduce a cooling tower efficiency multiplier $\lambda > 0$ and update Eqns.~\eqref{eqn:wue_fixedapproach} and \eqref{eqn:wue_fixedcold} as $ W_{direct}^{FixedApproach} = \lambda \cdot (- 0.0001896\cdot T_{w}^2+0.03095 \cdot T_{w}+0.4442)$ and $W_{direct}^{FixedColdWater} = \lambda \cdot ( 0.0005112\cdot T_w^2 - 0.04982\cdot T_w + 2.387)$. The value of $\lambda$ for a particular cooling system can be estimated by measuring its cooling load, water flow rate, range, and evaporation rate for a few different operational set points. For the rest of this paper, we consider $\lambda=1$ unless otherwise specified.}

\textbf{Estimation of indirect water.}
The electricity consumed from the power grid comes from various generation sources such as coal, oil, hydro, natural gas, nuclear, wind, and solar. These various electricity generation sources have a varying degree of water footprint associated with their electricity generation. However, once the electricity enters the power grid, it is difficult to separate the generation sources. Hence, the water footprint embedded in the electricity from the power grid is a mix of water footprints from multiple sources. Following prior literature and common practice \cite{Water_Electricity_EWIF_Water_Intensity_WorkingPaper_WorldResourcesInstitute_2020_reig2020guidance}, we calculate indirect water footprint using a weighted mix of water footprints from energy sources where each source's weight corresponds to the fraction of electricity generation it is responsible for. We use the following formula
\begin{equation}
    W_{indirect}(t) = \frac{\sum_k e_k(t) \cdot w_k}{\sum_k e_k(t)}
\end{equation}
where $e_k(t)$ is the electricity generation from energy source $k$ at time $t$ and $w_k$ is water footprint of energy source $k$.
\begin{figure}[t!]
    \centering
    \subfigure[Direct WUE]{\includegraphics[width=0.48\textwidth]{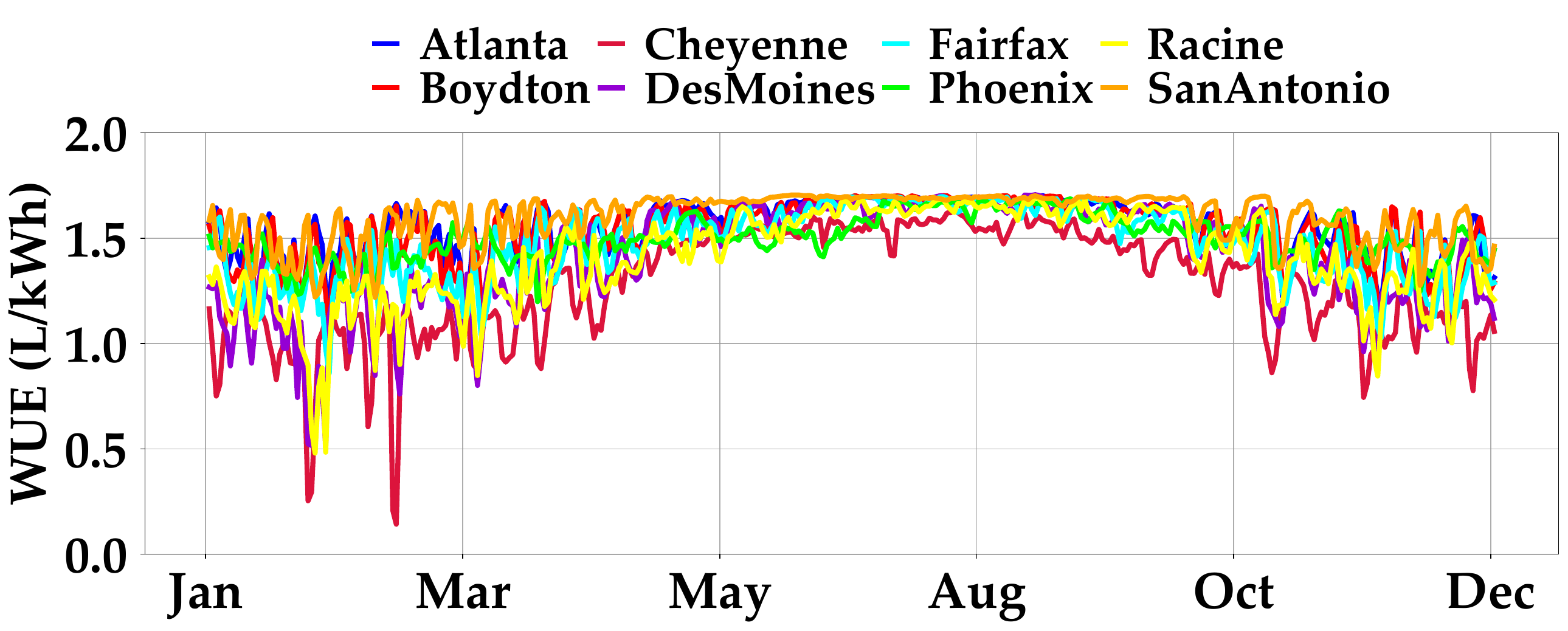}}
    \subfigure[Indirect WUE]{\includegraphics[width=0.48\textwidth]{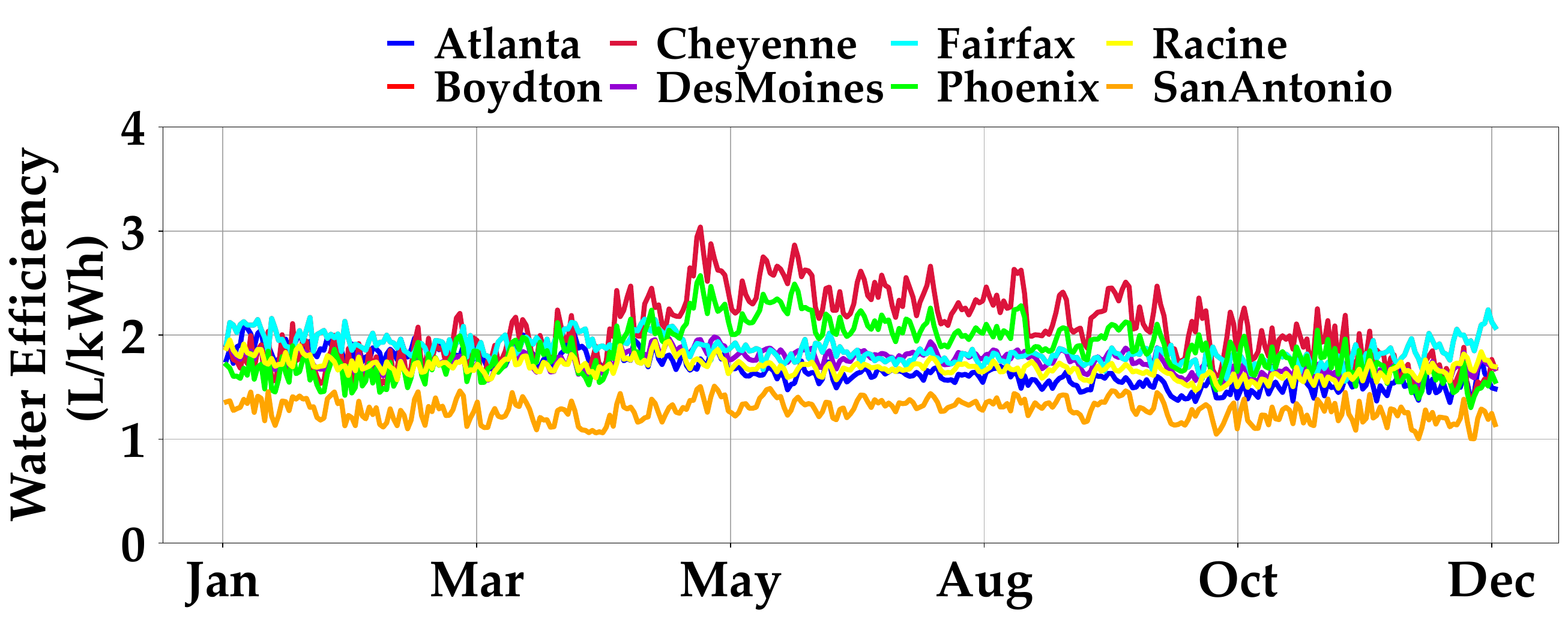}}
    \caption{Water efficiency across different U.S. locations in 2023.}
    \label{fig:timeseries_daily_avg}
\end{figure}

\subsection{Scope and Source of the Data}
We now provide some details of our dataset.

\textbf{Temporal resolution and duration.} In our dataset, we present five-year data from January \camera{2019} to December 2023. We collect our data at a temporal resolution of one hour. 

\textbf{Locations.} We incorporate \camera{58} major US cities with at least one city from each state \camera{(except Hawaii)}. 

\textbf{Weather data.}
To capture the temporal variation in direct WUE due to weather conditions, we collect the weather data from Weather Underground \cite{Weather_Data}. It offers location-wise hourly air temperature and relative humidity. We estimate the wet bulb temperature from the air temperature and relative humidity using the Stull formula \cite{stull2011wet}.

\textbf{Electricity data.}
We collect our hourly electricity data from EIA OpenData \cite{Water_EnergyData_EIA_Website}, which provides hourly electricity generation from different energy sources, as reported by the 76 different balancing authorities that operate in the U.S. Each balancing authority covers a certain geographical area. However, many of these balancing authorities interconnect with each other and, therefore, share the generation source of electricity. Hence, we follow the EPA eGrids zones mapping used for emission data \cite{Carbon_eGRID_US_EPA_Website}. eGrid divides the U.S. into 25 subregions and maps the balancing authorities to these subregions. We aggregate the balancing authority-level data into the eGrid subregion to determine electricity generation from different energy sources for a particular subregion.

\section{Preliminary Analysis of the Dataset}

\textbf{Time series visualization.}
Fig.~\ref{fig:timeseries_daily_avg} shows the time series daily average direct and indirect WUE for several U.S. cities. For the indirect WUE, we use the fixed approach model presented in Eqn.~\eqref{eqn:wue_fixedapproach}. We see a seasonal impact on the time series data where winter months show a larger day-to-day variation than summer months for direct WUE. The indirect WUE, on the other hand, does not reveal such seasonal impact prominently.

\begin{figure}[t!]
    \centering
\subfigure[Direct water]{    \includegraphics[width=0.48\linewidth]{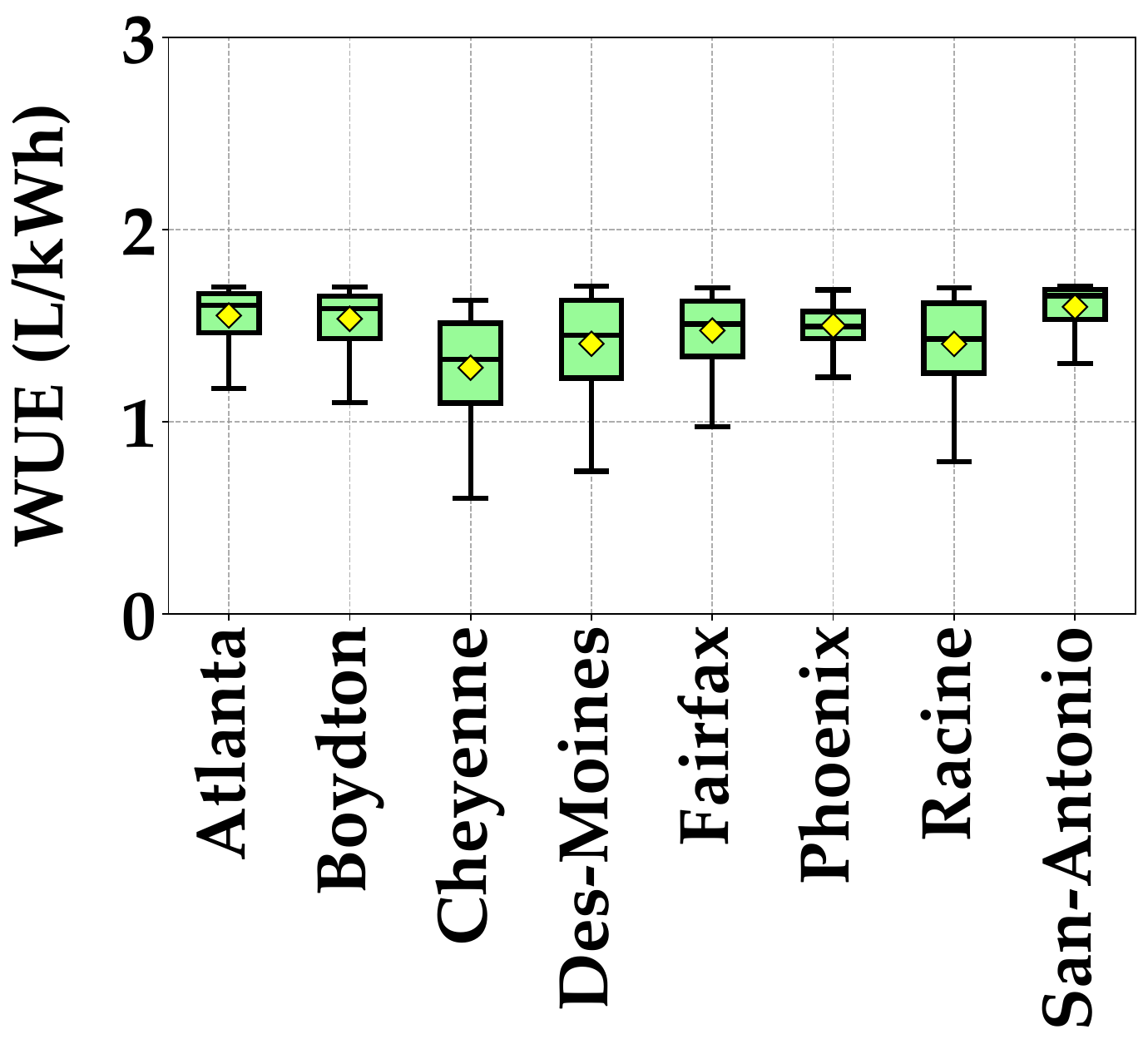}}
\subfigure[Indirect water]{    \includegraphics[width=0.48\linewidth]{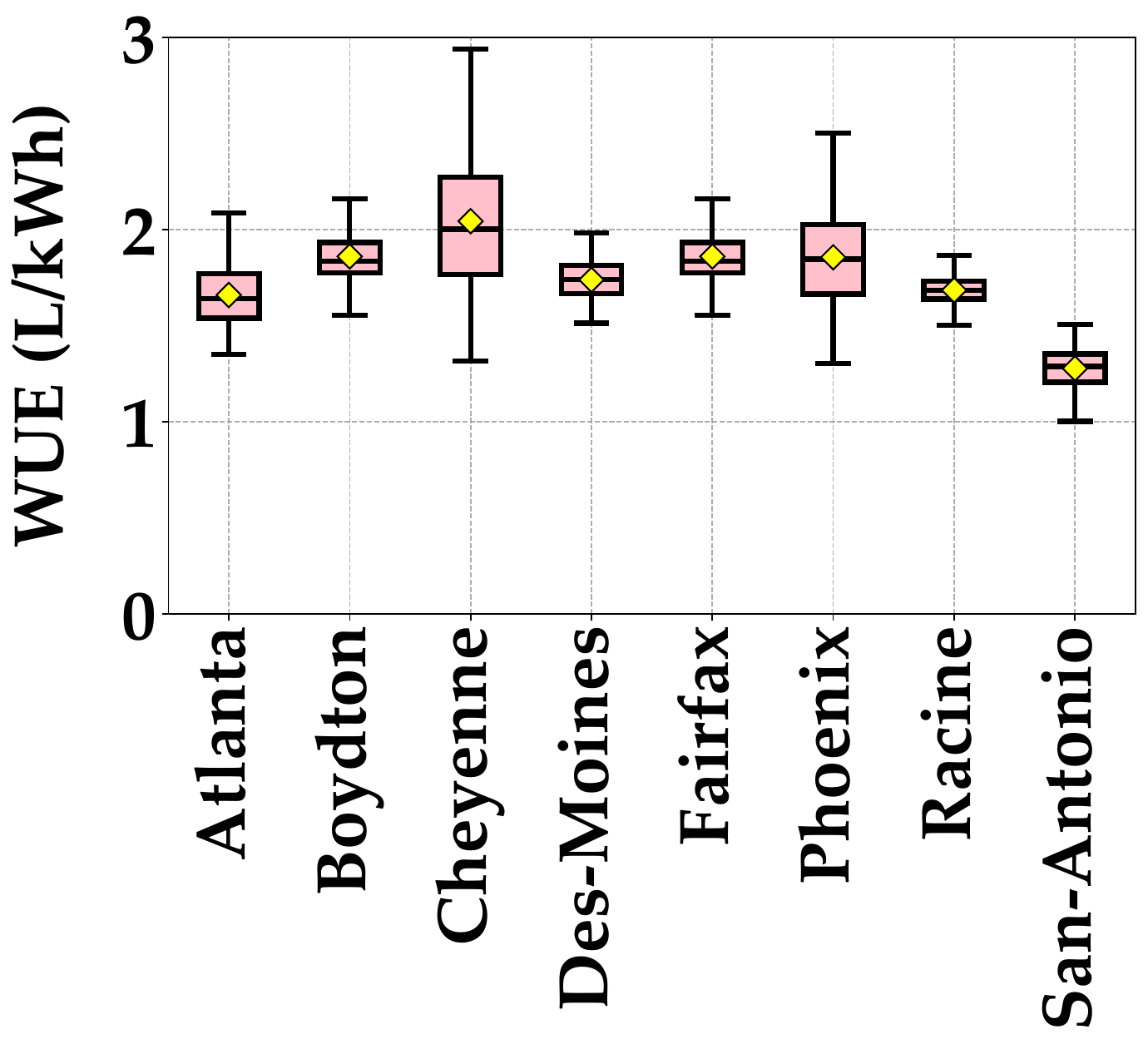}}
    \caption{Average WUE across different locations.}
    \label{fig:wue_location}
\end{figure}

\textbf{Spatial variation in WUE.}
Fig.~\ref{fig:wue_location} shows the direct and indirect WUE for several U.S. cities in 2023. This box plot extends from the first quartile (Q1) to the third quartile (Q3) of the data, with a black line at the median and the mean represented using a diamond marker. The whiskers extend from the box by 1.5x the inter-quartile range (IQR)
We see differences in both direct and indirect WUE due to differences in weather and the local mix of energy sources. We see particularly larger variations in indirect WUE across different locations. 

\begin{figure}[t!]
    \centering
\subfigure[Direct water]{    \includegraphics[width=0.48\linewidth]{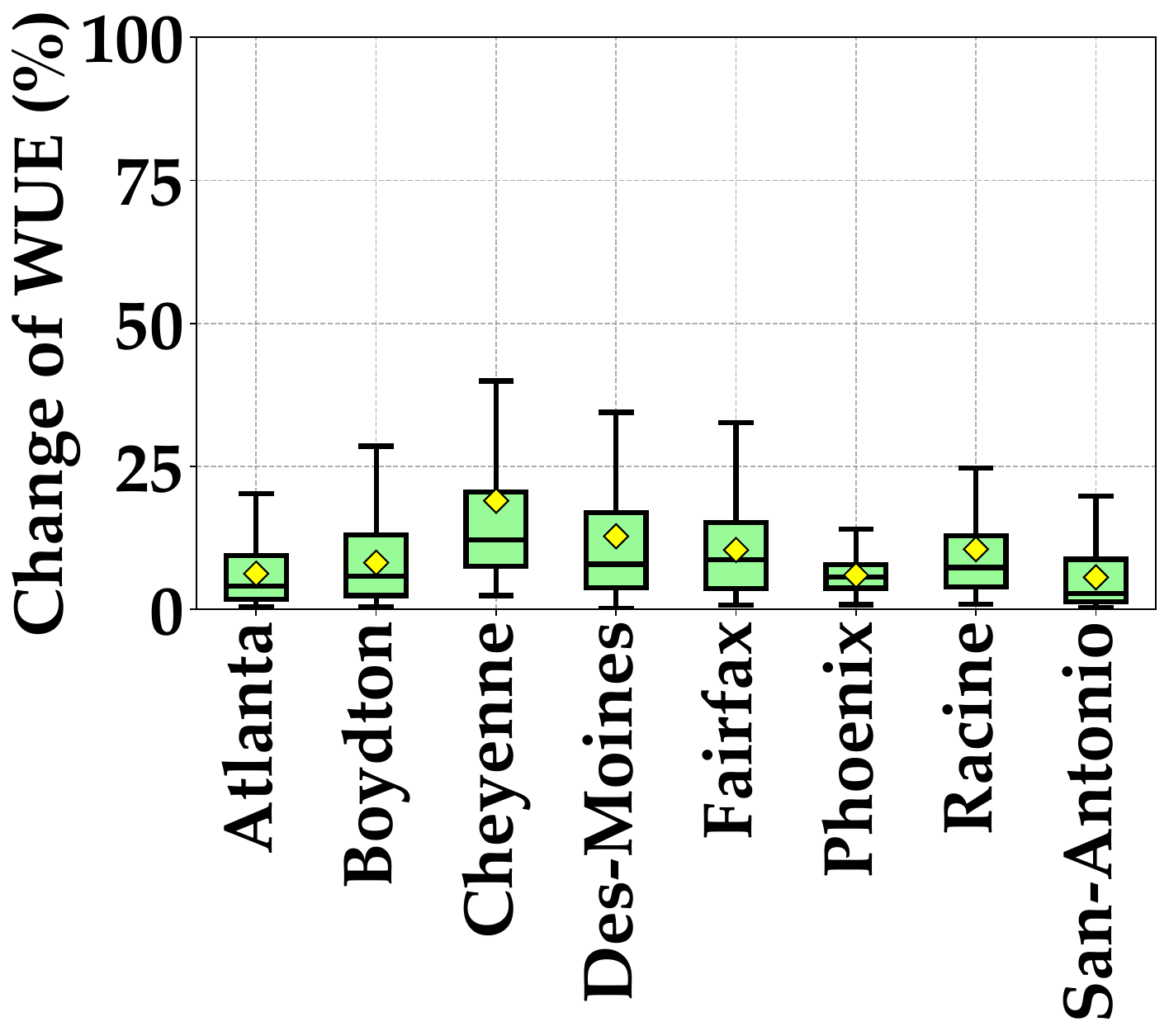}}
\subfigure[Indirect water]{  \includegraphics[width=0.48\linewidth]{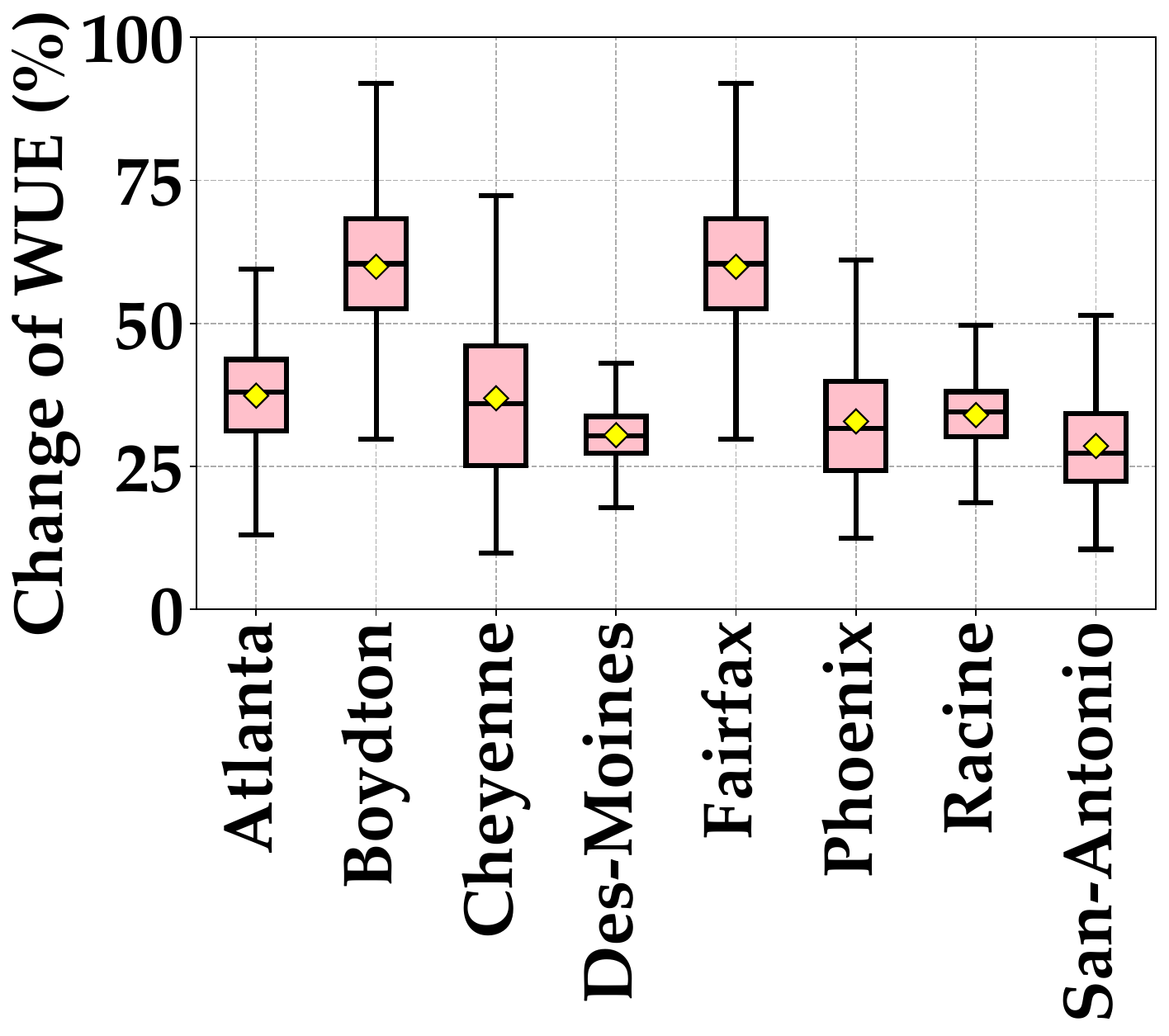}}
    \caption{Daily change in WUE.}
    \label{fig:wue_daily_variation}
\end{figure}

\textbf{Daily variations.}
In Fig.~\ref{fig:wue_daily_variation}, we show the maximum daily variations (difference between the maximum and minimum WUE over a single day) of WUEs in different locations. We see that, on average, the direct WUE varies as much as 20\% in places like Cheyenne. Meanwhile, the indirect WUE routinely varies more than 25\% across most locations shown in Fig.~\ref{fig:wue_daily_variation}.

\section{Example Applications}

We provide three example applications that may benefit from
 our dataset:
EV charging, building load management, and geographical load balancing in data centers.

\textbf{EV charging.}
EV charging  can benefit from integrating water footprint into their operation sustainability. EV charging mainly involves indirect water consumption due to its electricity consumption. EV charging activities also typically offer greater scheduling flexibilities \cite{alinia2019online}. Given the scheduling flexibility inherent in EV charging activities, implementing a water-aware charging schedule can optimize EV charging during periods of low indirect WUE, resulting in significant indirect water consumption savings. 

\textbf{Building load management.}
Commercial buildings commonly feature sizable centralized cooling systems employing cooling towers, resulting in both direct and indirect water consumption. 
Consequently, managing such buildings' cooling load and energy consumption can integrate the building's varying water efficiencies to reduce its overall water consumption.

\textbf{Geographical load balancing in data centers}
Water-aware scheduling and geographical load balancing offer substantial benefits to data centers and cloud applications \cite{liu2014greening,islam2016exploiting}. Data center workloads and cloud applications can exploit both temporal and spatial variation in the direct and indirect WUE. Unlike building load or EV charging, data center workloads can be moved around to more water-efficient locations. Many data center workloads, such as machine learning training, also offer great temporal scheduling flexibility and can be executed during water-efficient hours.

\section{Concluding Remarks}

In this paper, we introduced an hourly operational water efficiency dataset that captures the direct water consumption in the cooling system and indirect water embedded in the electricity generation. Additionally, we presented cooling system models capturing the impact of the weather conditions. We presented a preliminary analysis of our dataset highlighting the inherent temporal and spatial variation in direct and indirect WUEs. Furthermore, we discussed three potential applications that can benefit from utilizing our water efficiency dataset.

\begin{acks}
\camera{This work is supported in part by the US National Science Foundation under grants ECCS-2152357, CCF-2324915, and CCF-2324916.}
\end{acks}
\balance

\bibliographystyle{ieeetr}
\bibliography{main.bbl}

\balance

\end{document}